\documentclass[runningheads]{llncs}
\usepackage[T1]{fontenc}
\usepackage{graphicx}
\usepackage{multirow}
\usepackage{booktabs,array}
\usepackage[hang]{subfigure}

\begin{document}
\title{Replay to Remember: Continual Layer-Specific Fine-tuning for German Speech Recognition}
\titlerunning{Continual Layer-Specific Fine-tuning for German Speech Recognition}
%
\author{Theresa Pekarek Rosin \and 
Stefan Wermter} 
\authorrunning{T. Pekarek Rosin and S. Wermter}
%
\institute{Knowledge Technology, Department of Informatics, University of Hamburg,
Vogt-Koelln-Str. 30, 22527 Hamburg, Germany\\
\email{\{theresa.pekarek-rosin, stefan.wermter\}@uni-hamburg.de}\\
\url{www.knowledge-technology.info}}
\maketitle              
\begin{abstract}
While Automatic Speech Recognition (ASR) models have shown significant advances with the introduction of unsupervised or self-supervised training techniques, these improvements are still only limited to a subsection of languages and speakers. 
Transfer learning enables the adaptation of large-scale multilingual models to not only low-resource languages but also to more specific speaker groups. However, fine-tuning on data from new domains is usually accompanied by a decrease in performance on the original domain. Therefore, in our experiments, we examine how well the performance of large-scale ASR models can be approximated for smaller domains, with our own dataset of German Senior Voice Commands (SVC-de), and how much of the general speech recognition performance can be preserved by selectively freezing parts of the model during training. To further increase the robustness of the ASR model to vocabulary and speakers outside of the fine-tuned domain, we apply Experience Replay~\cite{Rolnick2018} for continual learning. By adding only a fraction of data from the original domain, we are able to reach Word-Error-Rates (WERs) below 5\% on the new domain, while stabilizing performance for general speech recognition at acceptable WERs.

\end{abstract}
\section{Introduction}
Automatic Speech Recognition (ASR) models have reached previously unseen state-of-the-art performance after the introduction of unsupervised and self-supervised pre-training methods from raw audio data, which allowed models to utilize a larger amount of speech data for training~\cite{Baevski2020,Conneau2020}. However, this has been accompanied by state-of-the-art models increasing in size and requiring thousands of hours of speech data to be trained properly. A recent example is Whisper~\cite{Radford2022}, which in its largest release contains 1550~M parameters and is trained on 680~000 hours of multilingual speech. 
\begin{figure}
\centering
\includegraphics[width=\textwidth]{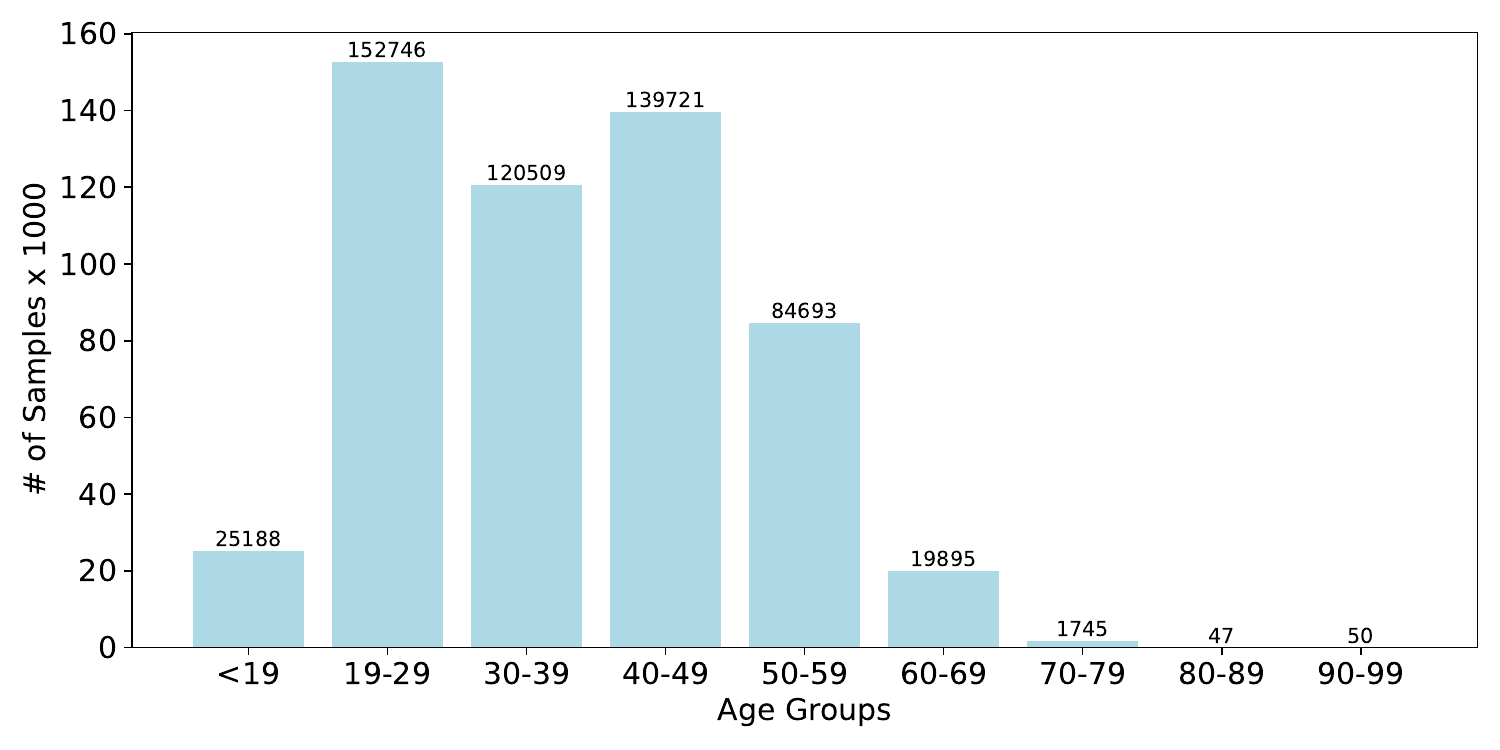}
\caption{The age distribution for Common Voice DE 10.0~\cite{Ardila2019}. As can be seen, of the labeled samples (ca. 70\%), the majority are between 19 and 59 years old. Older adults only constitute a fraction of the available samples.} \label{fig:cv_age}
\end{figure}

Fortunately, it is not necessary to train such a model from scratch for different languages and domains. Multilingual models like Whisper, or XLSR-53~\cite{Conneau2020} and its successor XLS-R~\cite{Babu2021}, generally perform better on low-resource languages than monolingual models trained from scratch, since similarities between languages can be leveraged. However, there is still improvement to be gained by fine-tuning for a specific language. For example, we observe a Word-Error-Rate (WER) of 15.2\% for Whisper-small~\cite{Radford2022} on Common Voice German 10.0 (CV-de)~\cite{Ardila2019} without any adaptation, still, through fine-tuning on additional hours of German speech this can be improved to 11.2\%~\cite{huang2022whisper-small-german}.

However, often a more specific adaptation for sub-groups or speakers is necessary, due to the fact that performance usually is much lower for speech that differs from the norm, e.g. due to accents, age, or speech disorders~\cite{MoroVelazquez2019,Ngueajio2022}.
This is due to the demographic distribution in most available datasets, where the majority of speakers are male, white, and middle-aged~\cite{Ngueajio2022}. As can be seen in Figure~\ref{fig:cv_age}, this issue transcends languages, as older age groups are similarly under-represented in CV-de~\cite{Ardila2019}, the most commonly used resource to train German speech recognition models. The same problem exists for the distribution of gender: of the subset labeled with additional demographic information in CV-de (ca. 70\%), female and diverse speakers only constitute 14\%.

To address this problem and thereby create more reliable ASR models, we can facilitate the knowledge contained in large-scale models, similar to how multilingual models can be utilized to improve ASR for low-resource languages. However, End-to-End ASR models also suffer from catastrophic forgetting~\cite{parisi2019}, even for within-language adaptation, which usually destroys the performance of general speech recognition~\cite{vandereeckt2022}. Therefore, a careful combination of transfer learning, i.e. leveraging the information contained in pre-trained models to facilitate learning on new domains, and continual learning, i.e. preventing the deterioration of performance on previously learned domains, is required.

We collect a dataset of German Senior Voice Commands (SVC-de), and compare the performance of Whisper~\cite{Radford2022}, XLSR-53~\cite{Conneau2020}, and XLS-R~\cite{Babu2021}, three state-of-the-art multilingual speech recognition models. We follow research for layer-specific fine-tuning~\cite{Huang2021,Shor2019} and examine how unfreezing different layer configurations influences the performance of the ASR model. Since domain adaptation usually leads to a decrease in performance on the original domain, we utilize Experience Replay (ER)~\cite{Rolnick2018} for continual learning to lessen the drop in performance for general speech recognition, and thereby increase the ASR model's robustness to out-of-domain vocabulary and speakers.

\section{Related Work}
\label{sec:relatedwork}
\subsection{Multilingual Speech Recognition}
The availability of pre-trained multilingual models in ASR has enabled transfer learning approaches for domains with limited data. This has been especially beneficial for improving speech recognition for non-standard speech and low-resource languages. 

XLSR-53~\cite{Conneau2020} and its successor XLS-R~\cite{Babu2021} are based on the wav2vec 2.0~\cite{Baevski2020} architecture and offer large-scale cross-lingual speech recognition. Pre-training in multiple languages, 53 for XLSR-53 and 128 for XLS-R, improves speech recognition across different languages since similarities between them are exploited during training.
Whisper~\cite{Radford2022} is a recent large-scale multilingual model, trained in an unsupervised manner for zero-shot cross-lingual speech recognition, speech translation, and language identification across 97 different languages with 680~000 hours of speech data. The underlying architecture is a simple encoder-decoder transformer~\cite{Vaswani2017}.

The results presented alongside these models show that multilingual ASR models usually perform better than monolingual models on low-resource languages. However, for languages where a large number of transcribed speech data is available, these models are outperformed by models utilizing supervised training~\cite{Babu2021,Radford2022}. This shows that it is beneficial to combine unsupervised pre-training with language- or domain-specific supervised fine-tuning.

\subsection{Layer-specific Fine-tuning}
While the transfer learning capabilities of large-scale speech recognition models have been demonstrated for multilingual~\cite{Babu2021,Conneau2020,Radford2022}, as well as monolingual adaptations~\cite{MacDonald2021,MoroVelazquez2019}, the question remains, if it is necessary to adapt the entire model during the fine-tuning process, especially for very specific or smaller domains. 

Shor et al.~\cite{Shor2019} fine-tune different layer combinations in Listen, Attend, and Spell (LAS) models~\cite{Chan2016} and RNN-T models~\cite{Graves2012} to find the subset of layers encoding the most information. For the LAS model, the best results are achieved through fine-tuning the entire model, but for the RNN-T model, 91\% of relative WER improvement is achieved by only fine-tuning the joint layer and the first layer of the encoder.

Similarly, Huang et al.~\cite{Huang2021} look different layer configurations of a Conformer-Transducer~\cite{Huang2020ConvTransformerTL} model in the context of efficient speaker adaptation. They observe that adaptation of the mid and bottom layers of the Conformer~\cite{Gulati2020} encoder offers a slight decrease in WER over adaptation of the top layers. 

Shrivasta et al.~\cite{Shrivastava2021} examine how much model performance depends on trained weights in the encoder and decoder of RNN-T~\cite{Graves2012} and Conformer models~\cite{Gulati2020}. They randomly initialized different parts of the model and confirmed that, while randomly initializing the encoder immediately hurt model performance, there was no significant difference in results for randomly initializing the decoder. 

While research on layer-specific fine-tuning has been mainly focused on performance approximations for new domains, the loss of performance for general speech recognition in monolingual layer-specific adaptations has not been examined in detail. The occurrence of catastrophic forgetting might be greatly dependent on the number of updated parameters, while the performance of attention-based models on the fine-tuned domain might not be. Therefore, we examine layer-specific fine-tuning for both domains, to see how much knowledge is gained for the new domain and lost for out-of-domain speech recognition in each configuration.

\subsection{Experience Replay}
Experience Replay (ER)~\cite{Rolnick2018} is a rehearsal-based continual learning (CL) method that aims to counteract catastrophic forgetting~\cite{parisi2019} by including a small fraction of data from the original domain in the training data for the new domain.
While CL for speech recognition is still relatively unexplored, ER has been utilized successfully for monolingual Dutch accent adaptation before~\cite{vandereeckt2022}. One advantage of rehearsal-based CL methods is that as long as data from the original domain is available or can be generated, the approaches can be used in a model-agnostic fashion. 

\begin{table}
\caption{An overview of the WER (\%) of our baseline models (\textit{-de} indicates fine-tuning on CV-de), evaluated on the test-split of Common Voice DE 10.0~\cite{Ardila2019} and our own German Senior Voice Commands (SVC-de) dataset.}\label{tab:wer_pt}
\centering
\renewcommand{\arraystretch}{1.4}
\begin{tabular}{l@{\hskip 0.15in}c@{\hskip 0.15in}c@{\hskip 0.15in}r}
\toprule
\multirow{2}{*}{\textbf{Model}} & \multirow{2}{*}{\textbf{CV-de 10.0}} & \multirow{2}{*}{\textbf{SVC-de}} &  \multirow{2}{*}{\textbf{\# of}}\\
& \textbf{test}& & \textbf{Params} \\
\midrule
XLSR-53-large-de~\cite{grosman2021xlsr53-large-german} &  12.8  & 18.4 &  315~M\\ 
XLS-R-1B-de~\cite{grosman2021xlsr-1b-german} & 11.6  & 20.0 &  962~M \\ 
XLS-R-300m-de~\cite{mcdowellxls-r300} & 22.8 & 31.3 & 315~M \\ 
Whisper-base-de & 20.4 & 25.7 & 74~M\\ 
Whisper-small-de~\cite{huang2022whisper-small-german} & 11.2 & 18.4 & 244~M\\ 
\bottomrule
\end{tabular}
\end{table}

\section{Experiments}
\label{sec:experiments}

\subsection{Data}
We fine-tune the models on the German Senior Voice Commands (SVC-de) dataset, a dataset we collected for the development of an ASR system for German senior citizens in the context of a home assistant system. The data has been collected with the approval of the Ethics Commission at the University of Hamburg.

SVC-de consists of short speech commands recorded by German speakers between the ages of 50 and 99. Overall 30 people (21 female, 9 male) recorded 52 sentences each with two microphones, for a total of 3h 9m, with approximately 6-7 minutes of audio data per speaker. The recorded sentences were manually cut and transcribed afterward to give a realistic estimation of the examined ASR models' performance. While most of the transcripts are close to the ground truth, some speakers deviated by repeating, skipping, or adding words, or by completely restructuring the requested sentence.

Common Voice DE (CV-de)~\cite{Ardila2019} is one of the largest and most utilized German speech datasets, and features a large variety of recording conditions and speakers due to the crowd-sourced nature of the collection. We utilize CV-de 10.0, which has 1~136 validated hours of audio from 16~944 different speakers and contains additional demographic data (e.g. age group, gender, accent) for about 70\% of the samples. 

\subsection{Base Models}
As can be seen in Table~\ref{tab:wer_pt}, the performance for German speech varies even for large-scale ASR models. While fine-tuning on data like CV-de improves the average performance for German speech, the improvement does not immediately translate to elderly speech. Additionally, a higher number of parameters does not seem to automatically lead to better performance. The performance of XLS-R-1B-de~\cite{Babu2021}, a model with approximately 1~B parameters is comparable to its predecessor XLSR-53-large~\cite{Conneau2020} and to Whisper-small~\cite{Radford2022}, with only 244~M parameters, after fine-tuning on CV-de. 

In our experiments, we utilize a selection of pre-trained models from the publically available checkpoints in Huggingface's\footnote{\url{https://huggingface.co/}} model repository. All models are approximately the same size and have been adapted to German speech with CV-de. We include a pre-trained version of XLS-R, with 300~M parameters~\cite{mcdowellxls-r300}, a pre-trained XLSR-53-large model~\cite{grosman2021xlsr53-large-german}, and a pre-trained Whisper-small model~\cite{huang2022whisper-small-german}. XLSR-53-large and XLS-R-300M both consist of 24 encoder layers and use character-based tokenization. Whisper-small consists of 12 encoder- and 12 decoder-layers and utilizes a byte-level BPE text tokenizer for an output vocabulary size of 51865. All models include punctuation to some degree, but to enable a fair comparison, we normalize the generated transcripts before the evaluation.

\subsection{Experiments}
In all our experiments, unless specifically stated otherwise, we train our models for five epochs with a batch size of 128 and AdamW~\cite{Loshchilov2017} optimizer. The learning rate is set to 3e-4 for XLS-R and XLSR-53, and to 3e-5 for Whisper. It decays linearly after a warm-up of 50 steps. 
We set the dropout for XLSR-53 and XLS-R to 0.1, and use mean CTC loss reduction. All hyperparameters were determined empirically, by comparing the behavior of the models during the layer-specific fine-tuning experiments.
We train our models on an NVIDIA A100 80G graphics card.

\subsubsection{Transfer Learning.}
The transfer learning capabilities of large pre-trained speech recognition models have been proven for adaptation to non-standard speech before~\cite{MacDonald2021,Shor2019}. Therefore, we establish a baseline by fine-tuning the entirety of our selected models on the SVC-de dataset. For XLSR-53 and XLS-R, we keep the feature extractor frozen.

Then, following similar approaches~\cite{Huang2021,Shor2019,Shrivastava2021}, we fine-tune different layer combinations to determine the most efficient subset for the adaptation of the model. 
Table~\ref{tab:layerconf} shows the layer configurations for our baseline models. Since XLSR-53-large~\cite{Conneau2020} and XLS-R-300m~\cite{Babu2021} share a network structure of 24 encoder layers, we can apply the same configurations to both models. While Whisper contains the same number of layers, the network structure is split into 12 encoder and 12 decoder layers. Therefore, we examine the layer configurations for the encoder and decoder in separate experiments and then apply them to both parts of the model simultaneously. For example, in the encoder-decoder fine-tuning scenario, we would apply the `first 6' configuration to both the encoder and the decoder, leading to a total of 12 adaptable layers.

We examine how much the performance differs between layer configurations for SVC-de and how much the performance for CV-de degrades due to domain adaptation. This should serve as an indicator as to which parts of the model are essential for the creation of general speech representations and therefore more sensitive to change, and which parts can be adapted for another domain without affecting the performance of the original dataset too drastically.
\begin{table}
\caption{The fine-tuning layer configurations for our baseline models. XLSR-53-large~\cite{Conneau2020} and XLS-R-300m~\cite{Babu2021} share the same number of encoder layers and therefore we can apply the layer configurations to both models. Due to the encoder-decoder architecture of Whisper~\cite{Radford2022}, we apply these configurations first to the 12 layers of the encoder and the 12 layers of the decoder separately, and then for both simultaneously.}\label{tab:layerconf}
\centering
\renewcommand{\arraystretch}{1.1}
\begin{tabular}{l@{\hskip 0.2in}l@{\hskip 0.15in}|@{\hskip 0.15in}l@{\hskip 0.2in}l}
\toprule
\multicolumn{2}{c}{\textbf{XLS-R \& XLSR-53}} & \multicolumn{2}{c}{\textbf{Whisper}}\\
\textbf{Name} & \textbf{Layer Configuration} & \textbf{Name} & \textbf{Layer Configuration} \\
\midrule
first 12 & [0, 1, ..., 10, 11] & first 6 & [0, 1, 2, 3, 4, 5]\\
last 12 & [12, 13, ..., 22, 23] & last 6 & [6, 7, 8, 9, 10, 11] \\
f4-i4-l4 & [0, ..., 3, 10, ..., 13, 20, ..., 23] & f1-i2-l1 & [0, 5, 6, 11]\\
f2-i2-l2 & [0, 1, 11, 12, 22, 23] & f2-i2-l2 & [0, 1, 5, 6, 10, 11] \\
last 6 & [18, 19, 20, 21, 22, 23] & last 3 & [9, 10, 11]\\
\bottomrule
\end{tabular}
\end{table}

\subsubsection{Continual Learning.}
To reduce the loss of knowledge regarding general speech recognition, we implement Experience Replay (ER)~\cite{Rolnick2018} for continual learning. However, instead of including a fixed number of samples from the original domain in each batch, we include either 10\% or 20\% of the original domain in the SVC-de training data spread out over all batches. We examine these data splits for the models with the best layer configurations, regarding their WER reduction on CV-de and their WER and convergence on SVC-de. We compare the performance between our best models with and without ER for both datasets.

\section{Results and Discussion}
\label{sec:results}
\subsection{Layer-Specific Fine-tuning}
\begin{figure}
\includegraphics[width=0.49\textwidth]{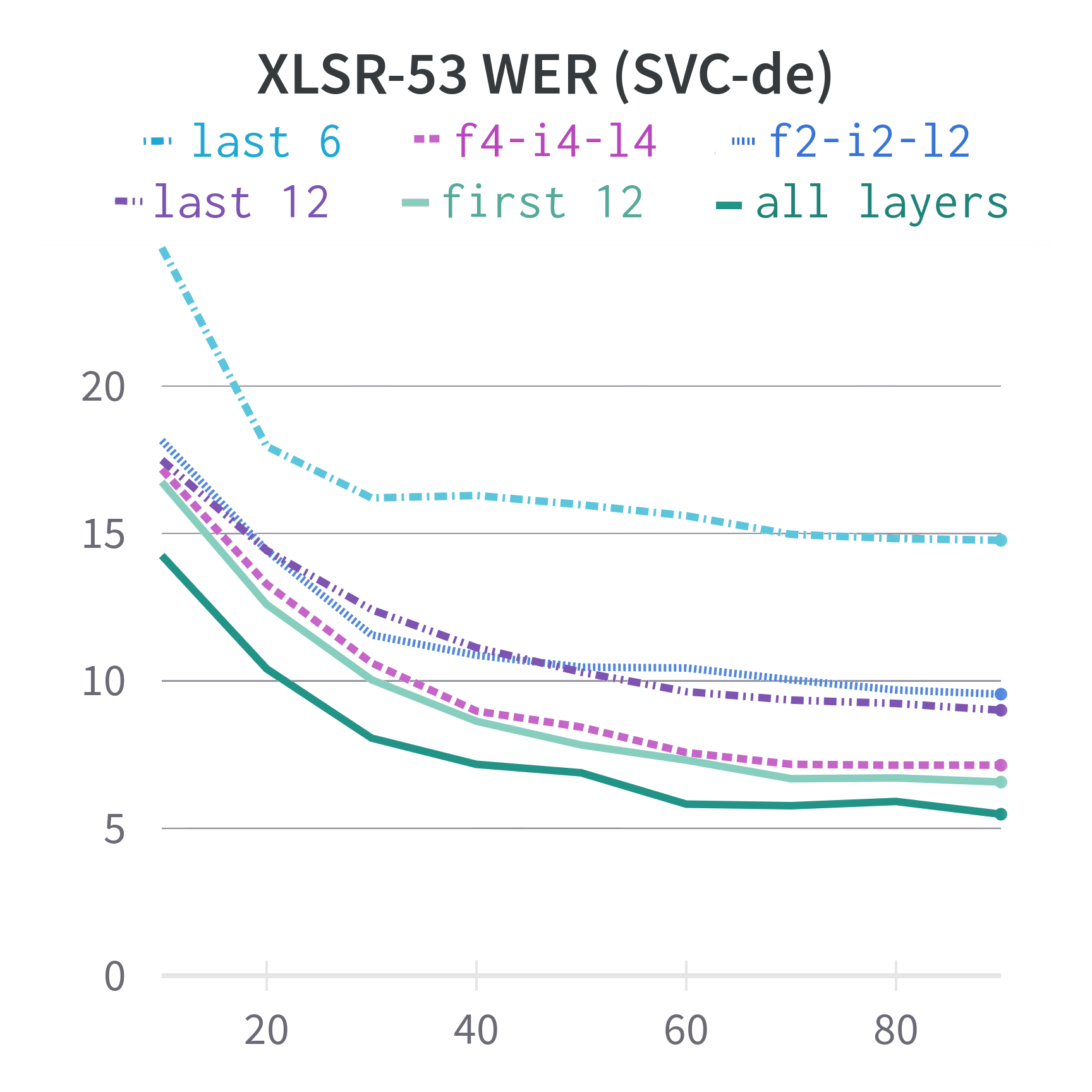}
\includegraphics[width=0.49\textwidth]{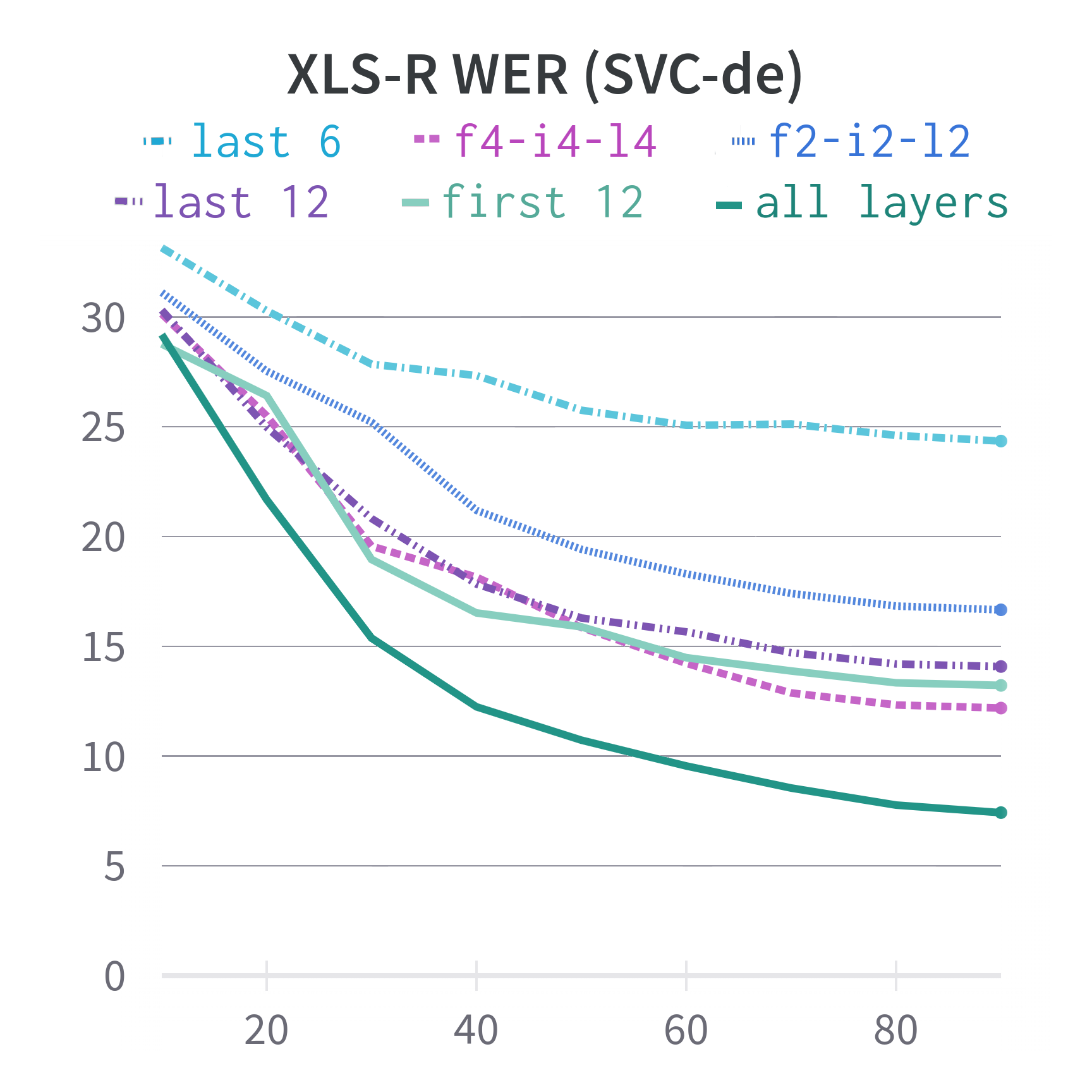}\\
\includegraphics[width=0.49\textwidth]{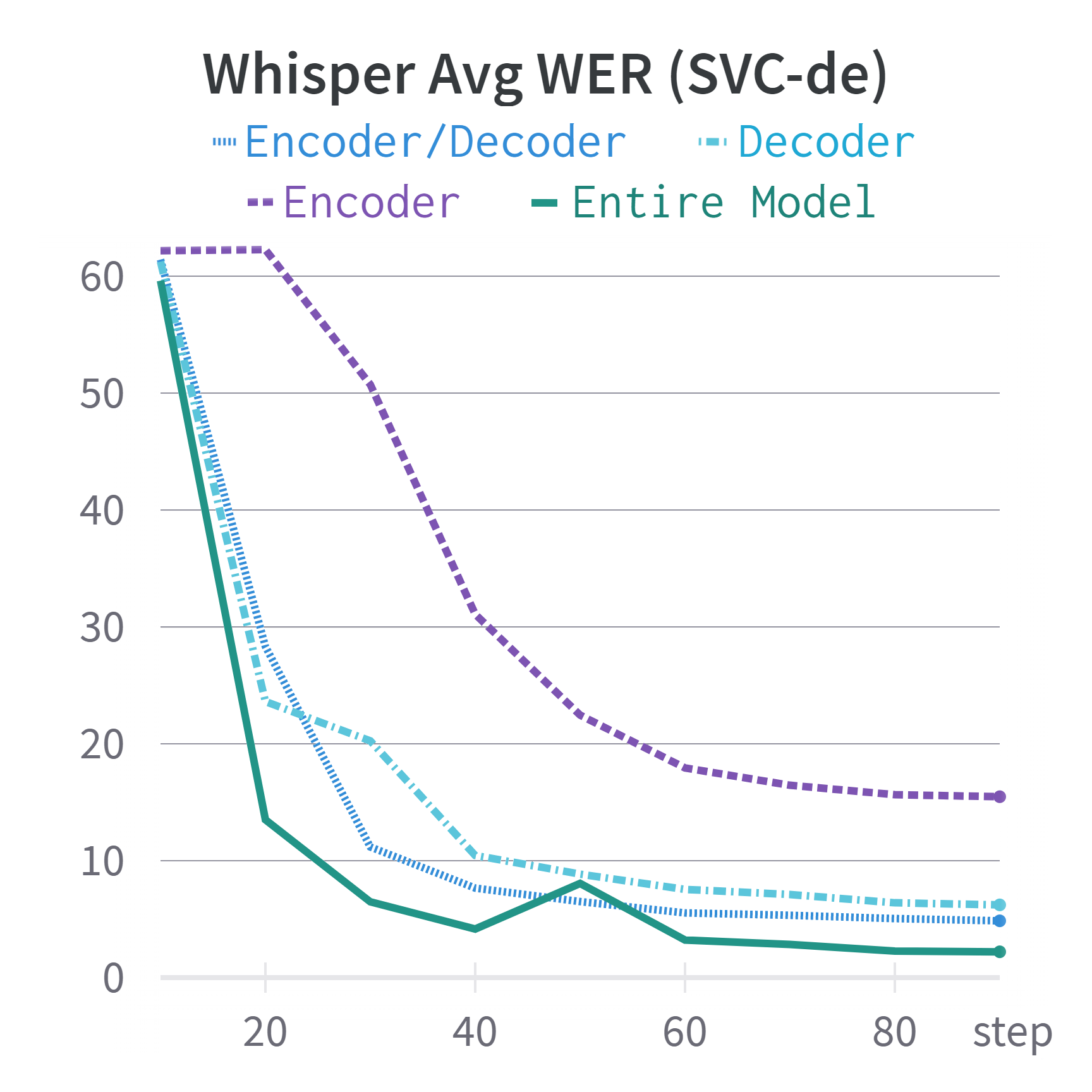}
\includegraphics[width=0.49\textwidth]{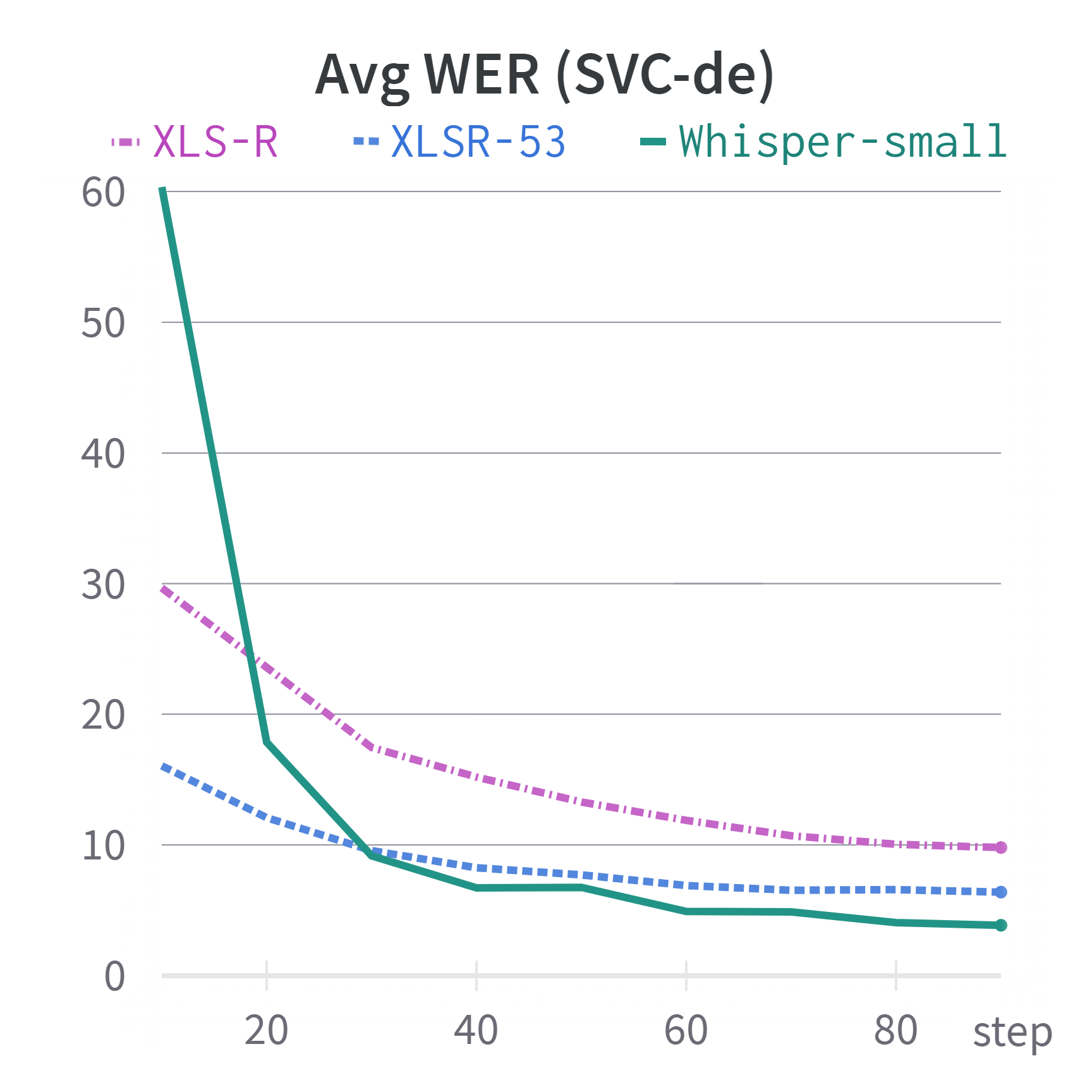}
\caption{The results of the layer-specific fine-tuning on SVC-de. For all models, the largest increase in performance can be observed after fine-tuning the entire model. However, for Whisper-small this performance can be approximated by layer configurations that only adapt the decoder or both model parts in unison. While XLSR-53 also offers a close approximation for some layer configurations, this is not the case with XLS-R. On average, Whisper's best layer configurations outperform both XLSR-53 and XLS-R after five epochs of training.}  \label{fig:ft_model_comp}
\end{figure}

As can be seen in Figure~\ref{fig:ft_model_comp}, fine-tuning the entire model generally leads to the best performance for all examined models. This aligns with the observations by Shor et al.~\cite{Shor2019} in their experiments with LAS. However, Whisper shows a clear difference in performance between layer configurations that adapt only the layers of the encoder or the decoder. Fine-tuning only the encoder layers leads to a final average WER of 15.5\%, which is an average increase of 13.3\% compared to the final WER obtained by fine-tuning the entire model (2.2\%). The encoder-decoder layer configurations reach an average WER of 4.8\% and the decoder configurations follow close behind with a final average WER of 6.2\% after five epochs.

The closest approximation of fine-tuning the entire Whisper model is obtained by fine-tuning the last six layers of the encoder and the decoder at the same time (WER: 3.1\%), followed closely by fine-tuning only the decoder (WER: 3.5\%). For XLSR-53, adapting only the first 12 layers (WER: 6.6\%) or configuration `f4-i4-l4' (WER: 7.1\%) offers a close approximation of the best model performance (WER: 5.5\%). Meanwhile, XLS-R shows the largest gap in WER between fine-tuning the entire model (WER: 7.4\%) and the next best `f4-i4-l4' configuration (WER: 12.2\%), but also the largest improvement on SVC-de, compared to its performance before the adaptation (Table~\ref{tab:wer_pt}).
However, Whisper outperforms both XLS-R and XLSR-53 on average after five epochs of training, despite an initial spike in WER on SVC-de.
\begin{figure}
\includegraphics[width=0.49\textwidth]{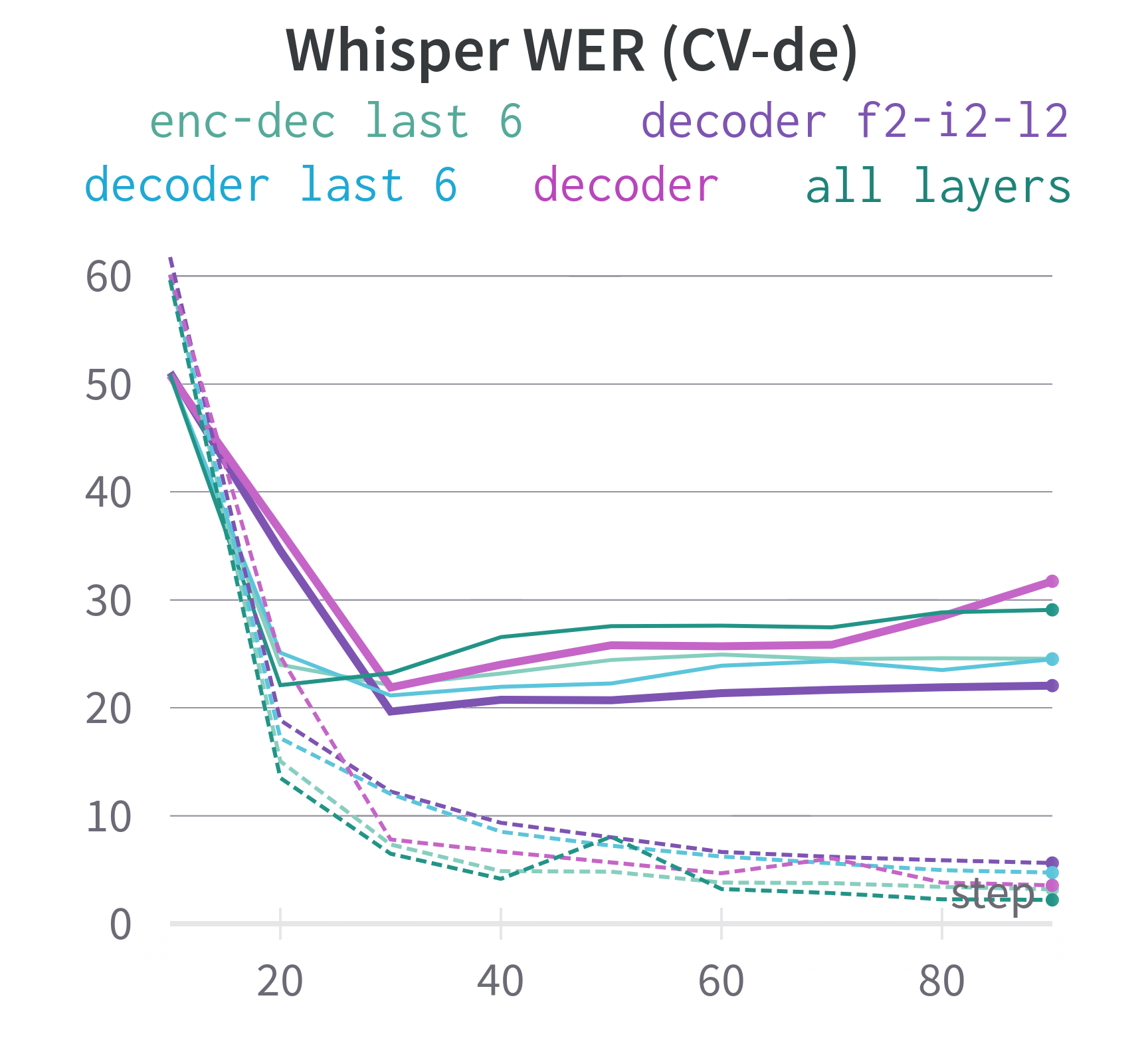}
\includegraphics[width=0.49\textwidth]{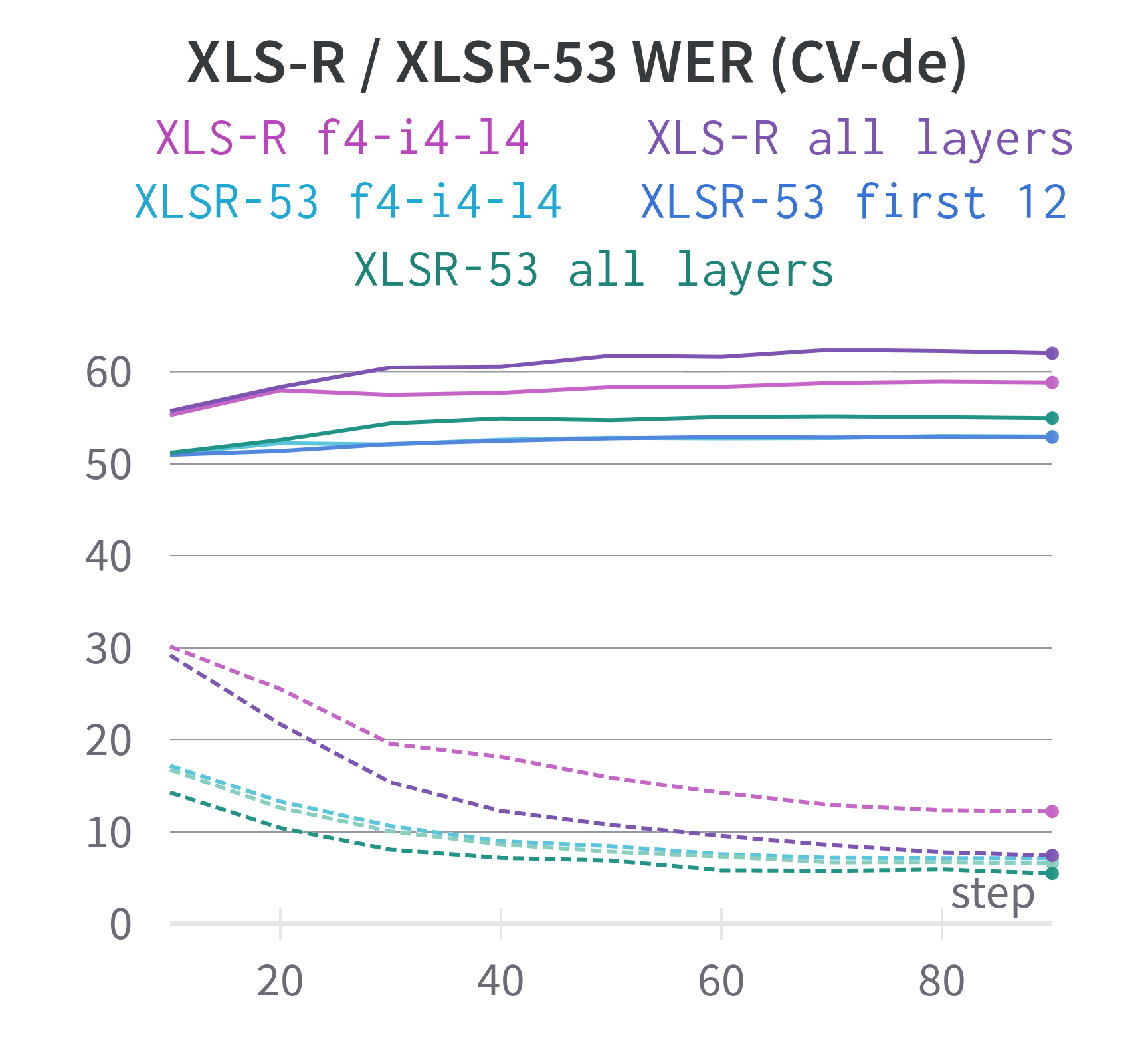}
\caption{The performance decay of CV-de during fine-tuning on SVC-de, measured in WER (lower is better, dashed lines indicate corresponding results on SVC-de). The most forgetting occurs for Whisper if the entire model is fine-tuned or all decoder layers are fine-tuned. For XLS-R and XLSR-53 the largest decay of performance happens (1) within the first 10 optimization steps and (2) when the entire model is fine-tuned on SVC-de. }  \label{fig:ft_cv_model_comp}
\end{figure}

As expected, the performance of CV-de deteriorates as a result of the fine-tuning process. Figure~\ref{fig:ft_cv_model_comp} shows a drastic increase in WER for all layer configurations and all examined models. However, the most forgetting occurs when the entire model is trained, and fine-tuning only a reduced number of layers generally leads to a lower WER for CV-de. This is especially interesting for cases, where adapting a smaller selection of layers is a close approximation of the original model performance. For example, fine-tuning only the last 6 layers of Whisper's encoder and decoder achieves a similar WER on SVC-de as adapting the entire model, with a difference of only 0.9\%. The WER on CV-de, however, is approximately 5\% lower for the smaller selection (24.5\%) compared to the entire model (29.1\%), which indicates that adapting only a smaller layer configuration is beneficial for preserving the performance of the original domain.

For XLS-R and XLSR-53 the behavior is similar, as most forgetting occurs when the entire model is fine-tuned. But, compared to Whisper, the WER on CV-de does not show any major changes after the first 10 optimization steps and is generally much higher. This is due to the selected learning rate. While a learning rate of 3e-3 leads to a better performance on SVC-de, the decay on CV-de is even more drastic and the learning process is overall more unstable. A learning rate of 3e-4 offered the best trade-off between performance on the new and the old domain, comparatively. 

\subsection{Experience Replay}
After applying ER during the fine-tuning process, we observe that ER with as little as 10\% of original data not only helps to stabilize Whisper's training on SVC-de but also diminishes the performance decrease on CV-de. As can be seen in Table~\ref{tab:wer_er_best}, fine-tuning only the last 6 layers of the encoder and the decoder leads to our best performance, with a final WER of 18.1\% on CV-de and 3.0\% on SVC-de, after five epochs of training. This is closely followed by adapting only the last 6 layers of the decoder with ER on 20\% of CV-de.
XLS-R and XLSR-53 also experience a WER reduction from ER, even though they do not reach the same level of performance as Whisper.

\begin{table}[t]
\caption{A comparison of our best models with and without Experience Replay (ER). While all models benefit from ER, a trade-off can be observed if we increase the percentage of samples from CV-de. Of all examined models, Whisper is the only one that can be stabilized at an acceptable WER for CV-de, while showing vast improvements for SVC-de.}\label{tab:wer_er_best}
\centering
\renewcommand{\arraystretch}{1.1}
\begin{tabular}{l@{\hskip 0.15in}l@{\hskip 0.15in}c@{\hskip 0.15in}c@{\hskip 0.15in}c}
\toprule
\textbf{Model} & \textbf{Layer Config} & \textbf{ER} & \textbf{SVC-de} & \textbf{CV-de 10.0} \\

& & (\%) & WER (\%) & WER (\%)\\
\midrule
XLSR-53         & all layers & -   &  \textbf{5.5}  &  55.0  \\ 
XLSR-53         & first 12 & -   &  6.6  &  52.9  \\ 
XLSR-53         & all layers & 10 &  \textbf{5.6}  &  42.7   \\
XLSR-53         & all layers & 20 &  5.9  &  \textbf{31.6}  \\
\midrule
XLS-R           & all layers & -  &  \textbf{7.4}  & 62.0   \\ 
XLS-R           & all layers & 10 &  8.1  & 53.5 \\
XLS-R           & all layers & 20 &  8.1  & \textbf{47.3} \\
\midrule
Whisper-small   & all layers     & -     &  \textbf{2.2} & 29.1 \\ 
Whisper-small   & enc/dec last 6 & -     &  3.2 & 24.5 \\ 
Whisper-small   & enc/dec last 6 & 10 &  \textbf{3.0} & 18.1 \\
Whisper-small   & dec last 6     & 10 &  5.0 & 17.6 \\
Whisper-small   & dec last 6     & 20 &  4.9 & \textbf{16.8} \\
\bottomrule
\end{tabular}
\end{table}

\section{Conclusion and Future Work}
\label{sec:page}
In this work, we demonstrate the effectiveness of combining layer-specific fine-tuning and continual learning to improve performance for under-represented speaker groups, while keeping the performance for general speech recognition from deteriorating in the process. Adapting smaller layer sub-groups for specific domains can, depending on the choice of model and configuration, approximate the performance of a model that has been fine-tuned in its entirety. Additionally, since fewer parameters are adapted during training the performance decay on the original domain is decreased. 
We show that utilizing Experience Replay (ER)~\cite{Rolnick2018} with only a small fraction of data from the original domain can lead to vast improvements in WER for the original, as well as minor improvements for the new domain.

Our best model is a pre-trained German Whisper-small architecture~\cite{huang2022whisper-small-german,Radford2022}, fine-tuned on SVC-de with 10\% ER, which reduces the WER for SVC-de from 18.4\% to 3.0\%. By adapting only the last six layers of the encoder and the decoder, we are able to stabilize the performance of CV-de at 18.1\% WER. By adding more data from the original domain, the WER on the original domain can be lowered further. However, we observe that at 20\% ER a trade-off starts to happen, where the performance on CV-de can only be improved with detriment to the performance of the new domain. 

While we utilize our own novel dataset of elderly German speech (SVC-de) in our experiments, the methods we use are model- and dataset-independent, which indicates that our approach could be applied to other domains (e.g. dialects) as well.
Additionally, since the vocabulary in SVC-de is limited, our approach promises more robustness for out-of-domain words and a larger variety of speakers than traditional fine-tuning approaches.

\subsubsection{Acknowledgements} The authors gratefully acknowledge support from the German BMWK (SIDIMO), the DFG (CML, LeCAREbot) and the European Commission (TRAIL, TERAIS). We would also like to thank Henri-Leon Kordt for helping with the post-processing of our German Senior Voice Commands dataset.
%
%
%
\bibliographystyle{splncs04}
\bibliography{literature}

\begin{thebibliography}{10}
\providecommand{\url}[1]{\texttt{#1}}
\providecommand{\urlprefix}{URL }
\providecommand{\doi}[1]{https://doi.org/#1}

\bibitem{Ardila2019}
Ardila, R., Branson, M., Davis, K., Kohler, M., Meyer, J., Henretty, M.,
  Morais, R., Saunders, L., Tyers, F., Weber, G.: {Common Voice: A
  Massively-Multilingual Speech Corpus}. In: Proceedings of the 12th Language
  Resources and Evaluation Conference. European Language Resources Association,
  Marseille, France (2020)

\bibitem{Babu2021}
Babu, A., Wang, C., Tjandra, A., Lakhotia, K., Xu, Q., Goyal, N., Singh, K.,
  von Platen, P., Saraf, Y., Pino, J.M., Baevski, A., Conneau, A., Auli, M.:
  {XLS-R: Self-supervised Cross-lingual Speech Representation Learning at
  Scale}. In: Proceedings of INTERSPEECH 2022. pp. 2278--2282. ISCA, Incheon,
  Korea (2022)

\bibitem{Baevski2020}
Baevski, A., Zhou, H., Mohamed, A., Auli, M.: {Wav2vec 2.0: A Framework for
  Self-Supervised Learning of Speech Representations}. In: Proceedings of the
  34th International Conference on Neural Information Processing Systems
  (NeurIPS). Curran Associates Inc., Vancouver, BC, Canada (2020)

\bibitem{Chan2016}
Chan, W., Jaitly, N., Le, Q., Vinyals, O.: {Listen, Attend and Spell: A Neural
  Network for Large Vocabulary Conversational Speech Recognition}. In:
  Proceedings of 2016 IEEE International Conference on Acoustics, Speech and
  Signal Processing (ICASSP). pp. 4960--4964. IEEE Press, Shanghai, China
  (2016)

\bibitem{Conneau2020}
Conneau, A., Baevski, A., Collobert, R., Mohamed, A., Auli, M.: {Unsupervised
  Cross-lingual Representation Learning for Speech Recognition}. In:
  Proceedings of INTERSPEECH 2021. pp. 2426--2430. ISCA, Brno, Czechia (2021)

\bibitem{Graves2012}
Graves, A.: {Sequence Transduction with Recurrent Neural Networks}. ICML 2012
  Workshop on Representation Learning  (2012)

\bibitem{grosman2021xlsr53-large-german}
Grosman, J.: {Fine-tuned {XLSR}-53 Large Model for Speech Recognition in
  {G}erman}.
  \url{https://huggingface.co/jonatasgrosman/wav2vec2-large-xlsr-53-german}
  (2021)

\bibitem{grosman2021xlsr-1b-german}
Grosman, J.: {Fine-tuned {XLS-R} 1{B} Model for Speech Recognition in
  {G}erman}.
  \url{https://huggingface.co/jonatasgrosman/wav2vec2-xls-r-1b-german} (2022)

\bibitem{Gulati2020}
Gulati, A., Qin, J., Chiu, C.C., Parmar, N., Zhang, Y., Yu, J., Han, W., Wang,
  S., Zhang, Z., Wu, Y., Pang, R.: {Conformer: Convolution-augmented
  Transformer for Speech Recognition}. In: Proceedings of INTERSPEECH 2020. pp.
  5036--5040. ISCA, Shanghai, China (2020)

\bibitem{huang2022whisper-small-german}
Huang, B.: {Fine-tuned {Whisper} Model for Speech Recognition in {G}erman}.
  \url{https://huggingface.co/bofenghuang/whisper-small-cv11-german} (2022)

\bibitem{Huang2020ConvTransformerTL}
Huang, W., Hu, W., Yeung, Y.T., Chen, X.: {Conv-Transformer Transducer: Low
  Latency, Low Frame Rate, Streamable End-to-End Speech Recognition}. In:
  Proceedings of INTERSPEECH 2020. pp. 5001--5005. ISCA, Shanghai, China (2020)

\bibitem{Huang2021}
Huang, Y., Ye, G., Li, J., Gong, Y.: {Rapid Speaker Adaptation for Conformer
  Transducer: Attention and Bias Are All You Need}. In: Proceedings of
  INTERSPEECH 2021. pp. 1309--1313. ISCA, Brno, Czechia (2021)

\bibitem{Loshchilov2017}
Loshchilov, I., Hutter, F.: {Decoupled Weight Decay Regularization}. In:
  Proceedings of 7th International Conference on Learning Representations
  (ICLR). New Orleans, LA, USA (2019)

\bibitem{MacDonald2021}
MacDonald, R.L., Jiang, P.P., Cattiau, J., Heywood, R., Cave, R., Seaver, K.,
  Ladewig, M., Tobin, J., Brenner, M.P., Nelson, P.C., Green, J.R., Tomanek,
  K.: {Disordered Speech Data Collection: Lessons Learned at 1 Million
  Utterances from Project Euphonia}. In: Proceedings of INTERSPEECH 2021. pp.
  3066--3070. ISCA, Brno, Czech Republic (2021)

\bibitem{mcdowellxls-r300}
McDowell, A.: Fine-tuned {XLS-R} 300{M} model for speech recognition in
  {G}erman.
  \url{https://huggingface.co/AndrewMcDowell/wav2vec2-xls-r-300m-german-de}
  (2022)

\bibitem{MoroVelazquez2019}
Moro-Velazquez, L., Cho, J., Watanabe, S., Hasegawa-Johnson, M.A., Scharenborg,
  O., Kim, H., Dehak, N.: {Study of the performance of automatic speech
  recognition systems in speakers with Parkinson's Disease}. In: Proceedings of
  INTERSPEECH 2019. pp. 3875--3879. ISCA, Graz, Austria (2019)

\bibitem{Ngueajio2022}
Ngueajio, M.K., Washington, G.: {Hey ASR System! Why Aren’t You More
  Inclusive? Automatic Speech Recognition Systems’ Bias and Proposed Bias
  Mitigation Techniques. A Literature Review}. In: Proceedings of 24th
  International Conference on Human-Computer Interaction (HCI 2022). p.
  421–440. Springer-Verlag (2022)

\bibitem{parisi2019}
Parisi, G.I., Kemker, R., Part, J.L., Kanan, C., Wermter, S.: {Continual
  Lifelong Learning with Neural Networks: A Review}. Neural Networks
  \textbf{113},  54--71 (2019)

\bibitem{Radford2022}
Radford, A., Kim, J.W., Xu, T., Brockman, G., McLeavey, C., Sutskever, I.:
  {Robust Speech Recognition via Large-Scale Weak Supervision}. ArXiv  (2022)

\bibitem{Rolnick2018}
Rolnick, D., Ahuja, A., Schwarz, J., Lillicrap, T., Wayne, G.: Experience
  replay for continual learning. In: Proceedings of the 33rd International
  Conference on Neural Information Processing Systems (NeurIPS). pp. 348--358.
  Curran Associates, Inc., Vancouver, BC, Canada (2019)

\bibitem{Shor2019}
Shor, J., Emanuel, D., Lang, O., Tuval, O., Brenner, M., Cattiau, J., Vieira,
  F., McNally, M., Charbonneau, T., Nollstadt, M., Hassidim, A., Matias, Y.:
  {Personalizing ASR for Dysarthric and Accented Speech with Limited Data}. In:
  Proceedings of INTERSPEECH 2019. pp. 784--788. ISCA, Graz, Austria (2019)

\bibitem{Shrivastava2021}
Shrivastava, H., Garg, A., Cao, Y., Zhang, Y., Sainath, T.N.: {Echo State
  Speech Recognition}. In: Proceedings of 2021 IEEE International Conference on
  Acoustics, Speech and Signal Processing (ICASSP). pp. 5669--5673. IEEE Press,
  Toronto, ON, Canada (2021)

\bibitem{vandereeckt2022}
Vander~Eeckt, S., Van~Hamme, H.: {Continual Learning for Monolingual End-to-End
  Automatic Speech Recognition}. In: Proceedings of 30th European Signal
  Processing Conference (EUSIPCO). pp. 459--463. IEEE Press, Belgrade, Serbia
  (2022)

\bibitem{Vaswani2017}
Vaswani, A., Shazeer, N., Parmar, N., Uszkoreit, J., Jones, L., Gomez, A.N.,
  Kaiser, L., Polosukhin, I.: {Attention is All you Need}. In: Proceedings of
  the 31st International Conference on Neural Information Processing Systems
  (NeurIPS). pp. 5998--6008. Curran Associates, Inc., Long Beach, CA, USA
  (2017)

\end{thebibliography}

\end{document}